\pdfoutput=1

\documentclass[11pt]{article}

\usepackage[]{EMNLP2023}

\usepackage{times}
\usepackage{latexsym}

\usepackage[T1]{fontenc}

\usepackage[utf8]{inputenc}

\usepackage{microtype}

\usepackage{inconsolata}
\usepackage{url}
\usepackage{booktabs}
\usepackage{multirow}
\usepackage{xcolor,colortbl}
\usepackage{graphicx}        
\usepackage{amssymb}

\definecolor{lightblue}{RGB}{173,221,215}
\definecolor{Gray}{gray}{0.85}
\newcolumntype{g}{>{\columncolor{Gray}}c}

\title{Limitations of LLMs for High Risk Domains \\ Despite Domain-Specific Instruction Tuning} 
\title{Yes, no, maybe? Assessing Instruction-Tuned GPT-like Models in High-Risk Domains }
\title{Yes, no, maybe? \\ Can We Use Instruction-Tuned GPT-like Models in High-Risk Domains? }
\title{Yes, No, Maybe? \\ Can We Use Large Language Models in High-Risk Domains? }
\title{Yes, No, Maybe -- \\ Can We Use Large Language Models in High-Risk Domains? }
\title{Walking a Tightrope -- \\  Evaluating Large Language Models in High-Risk Domains} 

\author{Chia-Chien Hung\textsuperscript{1}, Wiem Ben Rim\textsuperscript{1}, Lindsay Frost\textsuperscript{1}, \\\textbf{Lars Bruckner}\textsuperscript{2}, \textbf{Carolin Lawrence}\textsuperscript{1} \\
\textsuperscript{1}NEC Laboratories Europe, Heidelberg, Germany \\
\textsuperscript{2}NEC Europe Ltd, EU Public Affairs Office, Brussels, Belgium\\
  \texttt{\{Chia-Chien.Hung, Wiem.Ben-Rim, Carolin.Lawrence, Lindsay.Frost\}@neclab.eu}\\
  \texttt{Lars.Bruckner@emea.nec.com}}
  
\begin{document}
 \maketitle
\begin{abstract}
High-risk domains pose unique challenges that require language models to provide accurate and safe responses. Despite the great success of large language models (LLMs), such as ChatGPT and its variants, their performance in high-risk domains remains unclear. Our study delves into an in-depth analysis of the performance of instruction-tuned LLMs, focusing on factual accuracy and safety adherence. To comprehensively assess the capabilities of LLMs, we conduct experiments on six NLP datasets including question answering and summarization tasks within two high-risk domains: legal and medical. Further qualitative analysis highlights the existing limitations inherent in current LLMs when evaluating in high-risk domains. This underscores the essential nature of not only improving LLM capabilities but also prioritizing the refinement of domain-specific metrics, and embracing a more human-centric approach to enhance safety and factual reliability. Our findings advance the field toward the concerns of properly evaluating LLMs in high-risk domains, aiming to steer the adaptability of LLMs in fulfilling societal obligations and aligning with forthcoming regulations, such as the EU AI Act.

\end{abstract}

\section{Introduction}
Large language models (LLMs) have revolutionized how the world views NLP~\citep{wei2022emergent, kojima2022large}. Their astonishing performance on many tasks has led to an exponential increase in real-world applications of LLM-based technology. However, LLMs have a tendency to generate plausible but erroneous information, commonly referred to as hallucinations~\citep{ji-eta-al-2023-surveyhallucination}. This phenomenon proves to be particularly detrimental within high-risk domains, underscoring the importance of accurate and safe model outputs~\citep{nori2023capabilities}. 

\begin{figure}[t]
\centering
\includegraphics[trim={0.8cm 0cm 0cm 0cm},clip,width=1.0\linewidth]{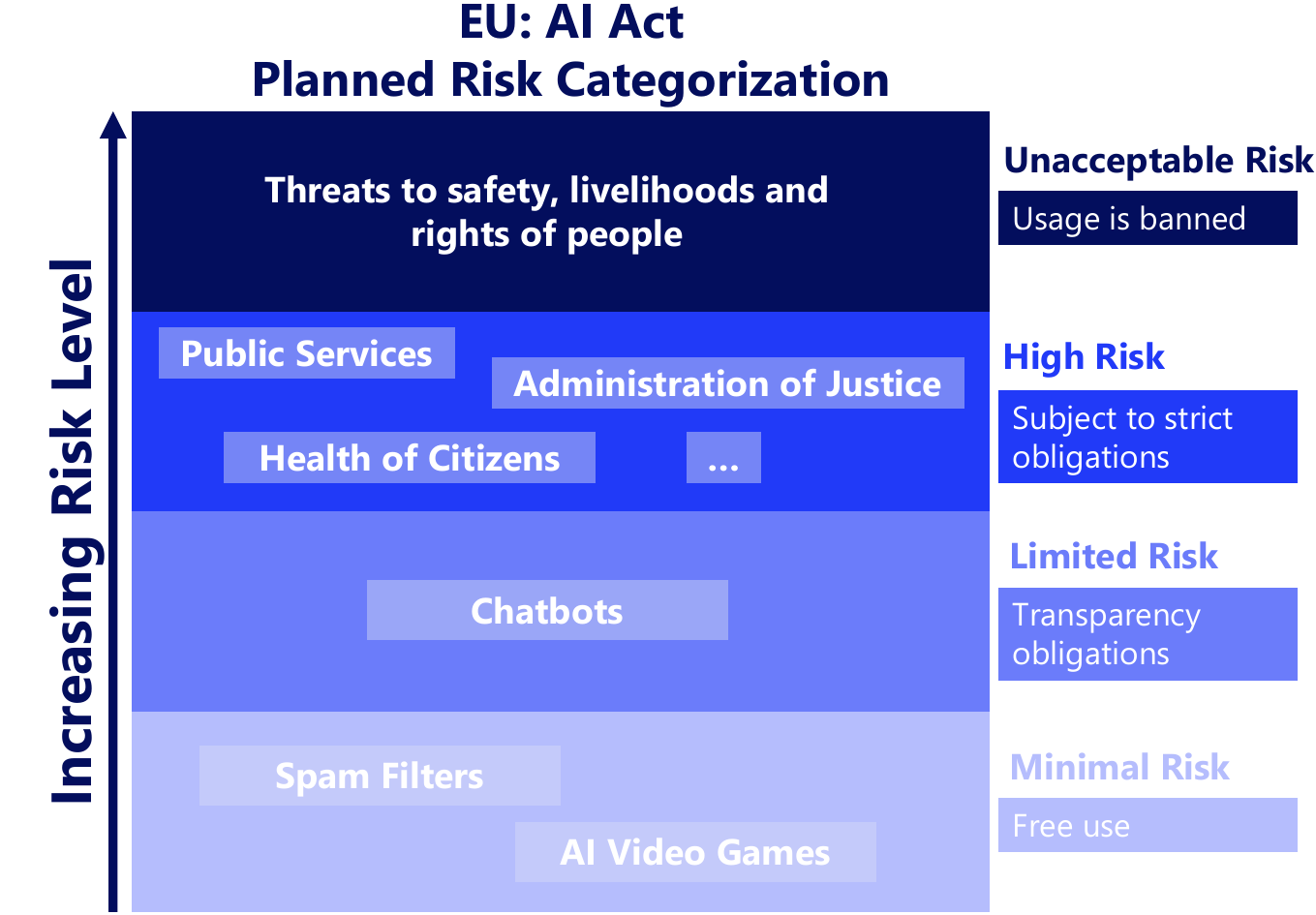}

\caption{The EU AI Act categorizes AI applications based on their associated risk levels. Although the Act is not yet finalized, it is expected that LLMs will fall into the high-risk category in specific domains, such as medical and legal.\footnotemark} 

\label{fig:eu_ai_act}
\end{figure}

In addition, with upcoming regulations, such as the EU AI Act~\cite{euai_proposal}, the necessity of properly analyzing and evaluating LLMs is further addressed. EU AI Act is expected to become the first law worldwide that regulates the deployment of AI  in the European Union, therefore, set a precedent for the rest of the world. According to the current draft, AI systems in high-risk domains, e.g. systems that have an impact on human life, will be subject to strict obligations, such as extensive testing and risk mitigation, prior to the system deployment (see Figure~\ref{fig:eu_ai_act}).
\footnotetext{Figure is based on \url{https://digital-strategy.ec.europa.eu/en/policies/regulatory-framework-ai}.}

In the era of LLMs, instruction-tuning~\citep{mishra-etal-2022-cross,wei-etal-2022-instructiontuning} has been proposed to efficiently solve various tasks like question answering (QA), summarization, and code generation~\citep{scialom-etal-2022-fine, wang2023codet5plus}. However, these models, trained on heterogeneous internet data, lack domain-specific knowledge crucial for accurate and reliable responses in high-risk domains, including up-to-date regulations, industry practices, and domain nuances~\citep{sallam2023chatgpt}. Furthermore, the quality of the training data is seldom quantified~\citep{zhou2023lima}. Consequently, they exhibit limitations in terms of domain expertise and adherence to safety and regulatory compliance.



In the study conducted by \citet{hupkes2022taxonomy}, a comprehensive perspective was introduced, advocating for the consideration of multiple facets in assessing generalization across diverse data distributions and scenarios. Building on the imperative of benchmarking generalization in the field of NLP and underscoring the importance of fairness in practical applications, our research delves into a specific yet pivotal dimension – \textit{how well can LLMs generalize effectively in high-risk domains?}   

Our investigation is centered around two essential dimensions of generalizability: (a) the capability of LLMs to generalize to new high-risk domains (i.e., general vs. high-risk domains) and new tasks (i.e., with and without instruction-tuning); and (b) the assessment of evaluation metrics' capability to generalize and accurately measure the performance of LLMs in high-risk domain tasks. Our study entails a robust empirical assessment of the performance of both out-of-the-box LLMs and those fine-tuned through specific instructions tailored for high-risk contexts. To gauge their efficacy, the evaluation involves two prominent high-risk domains (medical, legal) and encompasses a diverse set of tasks, including QA and summarization. 

\begin{figure}[t]
\centering
\includegraphics[width=1.0\linewidth]{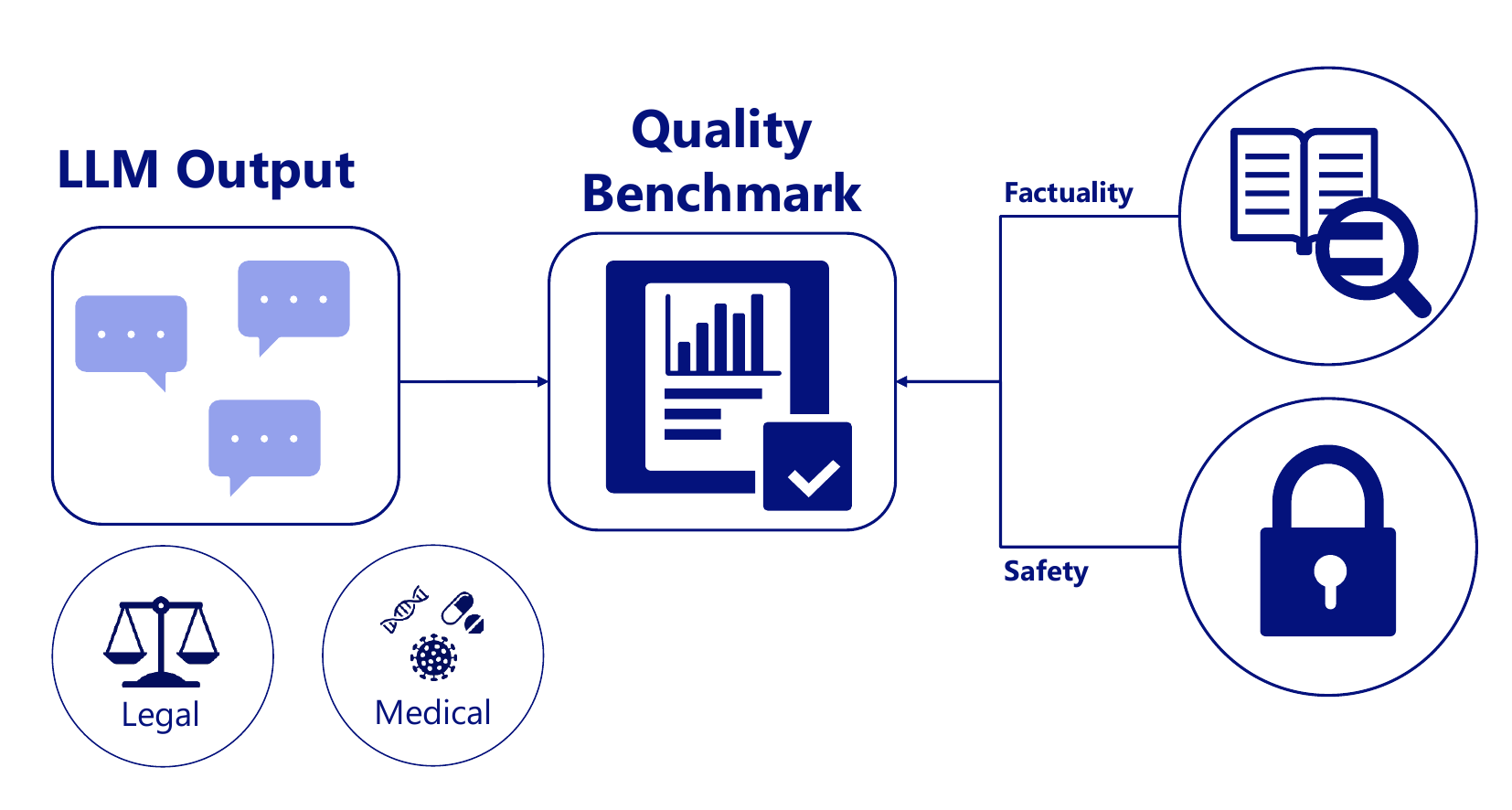}
\caption{Overview of the evaluation framework of evaluating LLMs in high-risk domains.  We evaluate how well LLMs with and without instruction-tuning perform in high-risk domains: legal and medical. The quality of the outputs is assessed using existing metrics to measure factuality and safety.} 
\label{fig:eval}
\end{figure}

We evaluate model outputs with regards to two key aspects, as depicted in Figure~\ref{fig:eval}: (1) \textit{factuality} -- are LLMs outputs factually correct for high-risk domains? (2) \textit{safety} -- do LLMs successfully avoid producing harmful outputs? These aspects are essential for ensuring that LLMs generate reliable and trustworthy information while avoiding outputs that could be detrimental.
To evaluate this, we employ existing metrics for \textit{factuality}~\citep{fabbri-etal-2022-qafacteval,zhong-etal-2022-towards} 
and \textit{safety}~\citep{Detoxify, dinan-etal-2022-safetykit} 
concerns. Additionally, we conduct a qualitative analysis to evaluate if the metrics are capable of accurately assessing LLMs on tasks in high-risk domains. 
Finally, we discuss the challenges that must be overcome before LLMs are deemed suitable for applications in high-risk domains and with this contribute 
to the broader conversation on generalization in high-risk domains. 

\paragraph{Contributions.}
Our contributions are summarized as follows: 
(i) We robustly evaluate the outputs of out-of-the-box and instruction-tuned LLMs in two high-risk domains on 6 datasets across QA and summarization tasks in terms of safety and factuality concerns; (ii) we demonstrate a qualitative investigation to identify shortcomings of existing metrics; (iii) 
we discuss open challenges that need to be solved in order to solidify trust to the generalization capability of LLMs in high-risk domains; (iv) we advocate for the need of human-centric NLP systems that are capable of giving the final control to human users in order to build trustworthy applications in high-risk domains.

\section{Domain-adaptive Instruction-tuning}




\label{subsec:instruction-tuning}
The emergence of GPT~\cite{radford-etal-2018-gpt} has led to a multitude of generative LLMs. One line of improving LLM performance has been proposed to increase the number of model parameters~\cite{chowdhery2022palm}. Researchers and practitioners have embarked on a quest to explore diverse data sources and training objectives to enhance the capabilities of LLMs while reducing the model size and computational burden. Another focus is leaning toward training smaller foundation models (e.g., GPT-J~\citep{gpt-j}, LLaMA~\citep{touvron-etal-2023-llama}, MPT~\citep{MosaicML2023Introducing}). The adoption of smaller foundation models enables researchers and practitioners to conduct more efficient investigations into novel methods, explore new domain-specific applications, and establish streamlined deployment efficiency. Crucially, the emphasis on smaller models is in accordance with the utilization of the instruction-tuning~\citep{mishra-etal-2022-cross} method, enabling efficient customization and adjustment of LLMs for particular domains or tasks~\citep{gpt4all, hu2023llm}.


In our experiments, we rely on a series of smaller size LLMs for efficiency and cost concerns, and effectively incorporate domain knowledge for high-risk domains via instruction-tuning. By leveraging explicit instructions during the training process, instruction-tuning has proved to enhance the model's ability for generalization~\citep{wei-etal-2022-instructiontuning} and domain adaptability~\citep{gupta-etal-2022-instructdial, wang2023codet5plus}. The domain-adaptive instruction-tuning approach explores the capability of how smaller models can effectively adapt to high-risk domains~\citep{yunxiang2023chatdoctor}. 

To efficiently incorporate domain knowledge, we employ QLoRA~\citep{dettmers2023qlora}, a method based on LoRA~\citep{hu-etal-2021-lora}, which compresses models using 4-bit quantization while maintaining performance parity. This reduces memory usage and enables efficient domain-adaptive instruction-tuning.

\section{Experimental Setup}
 
\paragraph{Instruction-tuning Data.}
\label{experiment:instruction_data}
To implement instruction-tuning, we collect in-domain datasets for legal and medical domains. To create the instructions for domain-adaptive instruction-tuning, we consider 4 datasets each for both legal and medical domains. An overview of the collected datasets is shown in Table~\ref{tab:instruction_data}. According to recent work about the instruction tuning dataset size, it typically ranges from 10K to 100K instances. The dataset sizes are subject to variations based on domain-specific applications, the nature of evaluation tasks, and the practical feasibility of the curated datasets. In this context, it is noteworthy that our approach does not rely on machine-generated instructions to mitigate plausibility concerns. Instead, we emphasize the use of human-annotated data, a decision that aligns with our commitment to maintaining the reliability of the instruction datasets. To ensure the efficacy of domain-adaptive instruction-tuning approach, we follow the steps from~\citep{wei-etal-2022-instructiontuning}, and construct templates for each of the datasets to form the final instructions. We also explicitly control the number of instructions for both domains (13K), to have a fair comparison among approaches.  Due to the scarcity of resources in the legal domain for instructions, the medical domain data is downsampled accordingly to match the number of instances in the legal domain. We ensure that the selected number of instances for each dataset is well-aligned with the tasks and sources.

\paragraph{Evaluation Tasks.}
We focus on two high-risk domains (legal and medical), aligned with EU AI Act domain categorization (see Figure~\ref{fig:eu_ai_act}), and evaluate 6 datasets across QA and summarization (SUM) tasks.  The tasks include \textit{multiple-choice QA}~\citep{zheng-etal-2021-casehold}, \textit{free-form QA}~\citep{li-etal-2022-parameter, yunxiang2023chatdoctor}, \textit{reasoning QA}~\citep{jin-etal-2019-pubmedqa}, and \textit{long document summarization}~\citep{kornilova-eidelman-2019-billsum,wallace-etal-2020-rctsum}. Table~\ref{tab:task_data} displays an overview of the high-risk domain task datasets. We provide example excerpts and templates designed for each task in Appendix~\ref{appendix:instruction_template}.

\setlength{\tabcolsep}{3.5pt}
\begin{table}[t]
\centering
\scriptsize{
\begin{tabular}{llrc}
\toprule
\textbf{Domain}                     & \textbf{Dataset}                                     & \textbf{Size}     & \textbf{License\dag}        \\ \midrule
\multirow{4}{*}{Legal}  & BillSum~\citep{kornilova-eidelman-2019-billsum}  &  88  &  CC0-1.0 \\ 
& CaseHold~\citep{zheng-etal-2021-casehold} & 2,458 &  CC-BY-SA \\ 
& LegalAdviceReddit~\citep{li-etal-2022-parameter} & 9,984 & CC-BY-SA \\ 
 & LawStackExchange~\citep{li-etal-2022-parameter} & 513  & CC-BY-SA \\ 
\midrule
 \multirow{4}{*}{\shortstack[l]{Medical}} & PubMedQA~\citep{jin-etal-2019-pubmedqa} & 513  &  MIT  \\
 & RCTSum~\citep{wallace-etal-2020-rctsum} & 151  &  Apache-2.0  \\
  & MedQA~\citep{jin2020disease} & 2,458  &  MIT  \\
 & HealthCareMagic~\citep{yunxiang2023chatdoctor}   &  10,000  & Apache-2.0 \\
\bottomrule
\end{tabular}%
}
\caption{Overview of the datasets utilized for instruction-tuning for high-risk domains (legal, medical). The size of the in-domain data and the \textit{commercial} applicability based on the license are reported. \dag License: Creative Commons Zero (cc0), Creative Commons Attribution Share-Alike (CC-BY-SA).} 
\label{tab:instruction_data}
\end{table}

%
\setlength{\tabcolsep}{1.6pt}
\begin{table}[t]
\centering
\scriptsize{
\begin{tabular}{llcrcc}
\toprule
\textbf{Domain}                     & \textbf{Dataset}                                      & \textbf{Task}        & \textbf{Size}   & \textbf{License}     \\ \midrule
\multirow{3}{*}{Legal} 
& BillSum~\citep{kornilova-eidelman-2019-billsum}               & SUM & 100            & cc0-1.0  \\ 
& CaseHold~\citep{zheng-etal-2021-casehold} & QA & 1000 & Apache-2.0 \\ 
 & LawStackExchange~\citep{li-etal-2022-parameter} & QA & 989 & CC-BY-SA\\\midrule 
\multirow{3}{*}{\shortstack[l]{Medical}}  &  PubMedQA~\citep{jin-etal-2019-pubmedqa}                 &   QA     &  250     & MIT \\
  &    RCTSum~\citep{wallace-etal-2020-rctsum}               &   SUM     &   100    & Apache-2.0 \\
  & iCliniq~\citep{yunxiang2023chatdoctor}   &  QA & 1000 & Apache-2.0 \\
                         \bottomrule
\end{tabular}%
}
\caption{Overview of the evaluation datasets for high-risk domains (legal, medical). For each domain, we report the task type, dataset size, and license. All the selected task datasets are applicable for \textit{commercial} usage.}
\label{tab:task_data}
\end{table}

\paragraph{Evaluation Metrics.}
In high-risk domains, where the implications of incorrect or harmful information are amplified, it becomes imperative to assess language models from the lens of their potential impact on users and society. The selection of \textit{factuality} and \textit{safety} as evaluation metrics is rooted in the following considerations:
(1) \textit{Factuality} is considered as the ability of LLMs to provide factual and precise responses. Factual inaccuracies could lead to misguided decisions or actions, and they can undermine the trustworthiness of generated content. By evaluating factuality, we seek to ensure that the responses of LLMs align with accurate information, which is of utmost importance in high-risk applications. Two metrics are considered and have been shown to align with human judgments: QAFactEval \cite{fabbri-etal-2022-qafacteval}, which measures fine-grained overlap of the generated text against the ground truth, and UniEval~\citep{zhong-etal-2022-towards}, which computes over several dimensions, namely coherence, consistency, fluency, and relevance. (2) \textit{Safety} is defined as the degree of insensibility and responsibility in the generated content that is safe, unbiased, and reliable. High-risk domains often involve sensitive topics, legal regulations, and ethical considerations, thus ensuring safety in the generated contents mitigates the potential of unintended consequences, such as perpetuating harmful stereotypes or generating discriminatory content~\citep{kaddour2023challenges}. Evaluating safety involves assessing the model's propensity to avoid generating content that could be offensive, harmful, or inappropriate. We consider Detoxify~\citep{Detoxify} and SafetyKit~\citep{dinan-etal-2022-safetykit}, which measure a model's tendencies to agree to offensive content or give the user false impressions of its capabilities as well as other safety concerns. 
Although our primary focus is on ensuring factuality and safety, it is essential to underscore the significance of other critical factors, such as \textit{robustness}~\citep{zhu2023promptbench}, that are also vital for evaluating LLMs. While acknowledging the broader spectrum of evaluation dimensions that warrant attention in comprehensive assessments of LLMs, our emphasis on \textit{factuality} and \textit{safety} is prioritized by the pressing and tangible concerns related to misinformation and potential harm in high-risk domains. Overall evaluation is aligned with AuditNLG\footnote{\url{https://github.com/salesforce/AuditNLG}} library.


\paragraph{Evaluation Card.}
Inspired by the generalization taxonomy introduced by~\citet{hupkes2022taxonomy} to characterize and gain insights into the field of generalization research in NLP, it comprises the following key dimensions for evaluation: (1) \textit{motivation (practical)}: we assess the generalization capabilities of models with the objective to be deployed for real-world high-risk domain tasks; (2) \textit{generalization type (cross-domain, cross-task)}: we investigate how effectively models generalize across different domains and tasks; (3) \textit{shift locus (pretrain-train, pretrain-test)} and \textit{shift type (label shift)}: the experimental results are compared with LLMs instruction-tuned on domain instructions and the ones without; and (4) \textit{shift source (naturally shift)}: we only consider human-annotated data to mitigate plausibility concerns (see \S\ref{experiment:instruction_data}). We summarize the generalizability of our proposed methods in Table~\ref{tab:eval_card}.

\definecolor{Mycolor2}{HTML}{00F9DE}
\newcommand{\expone}{$\checkmark$}
\setlength{\tabcolsep}{2.7pt}
\renewcommand{\aboverulesep}{0.3pt}
\renewcommand{\belowrulesep}{0.3pt}
\begin{table}[t]
\centering
\tiny{
\begin{tabular}{@{}clllllllllll@{}} 
\toprule
\rowcolor{lightblue}
\multicolumn{12}{c}{\textbf{Motivation}}\\
\multicolumn{3}{c}{\textit{Practical}} & \multicolumn{3}{c}{\textit{Cognitive}} & \multicolumn{3}{c}{\textit{Intrinsic}} & \multicolumn{3}{c}{\textit{Fairness}} \\
\multicolumn{3}{c}{\expone}                   & \multicolumn{3}{l}{}                   & \multicolumn{3}{l}{}                   & \multicolumn{3}{c}{}                  \\
\rowcolor{lightblue}  
\multicolumn{12}{c}{\textbf{Generalization type}}                                                                                                                \\
\multicolumn{2}{l}{\textit{Compositional}} &
  \multicolumn{2}{l}{\textit{Structural}} &
  \multicolumn{2}{l}{\textit{Cross Task}} &
  \multicolumn{2}{l}{\textit{Cross Language}} &
  \multicolumn{2}{l}{\textit{Cross Domain}} &
  \multicolumn{2}{l}{\textit{Robustness}} \\
\multicolumn{2}{l}{}      & \multicolumn{2}{l}{}     & \multicolumn{2}{c}{\expone}     & \multicolumn{2}{l}{}      & \multicolumn{2}{c}{\expone}     & \multicolumn{2}{c}{}    \\
\rowcolor{lightblue}
\multicolumn{12}{c}{\textbf{Shift locus}}                                                                                                                        \\
\multicolumn{3}{c}{\textit{Train--test}} &
  \multicolumn{3}{c}{\textit{Finetune train--test}} &
  \multicolumn{3}{c}{\textit{Pretrain--train}} &
  \multicolumn{3}{c}{\textit{Pretrain--test}} \\
\multicolumn{3}{l}{}                   & \multicolumn{3}{l}{}                   & \multicolumn{3}{c}{\expone}                   & \multicolumn{3}{c}{\expone}   \\
\rowcolor{lightblue}  
\multicolumn{12}{c}{\textbf{Shift type}}                                                                                                                         \\
\multicolumn{3}{c}{\textit{Covariate}} & \multicolumn{3}{c}{\textit{Label}}     & \multicolumn{2}{c}{\textit{Assumed}}      & \multicolumn{2}{c}{\textit{Full}} & \multicolumn{2}{c}{\textit{Multiple}}  \\
\multicolumn{3}{c}{}                   & \multicolumn{3}{c}{\expone}    &              \multicolumn{2}{c}{}  & \multicolumn{2}{c}{}                   & \multicolumn{2}{c}{}                  \\
\rowcolor{lightblue}
\multicolumn{12}{c}{\textbf{Shift source}}                                                                                                                       \\
\multicolumn{3}{c}{\textit{Naturally shift}} &
  \multicolumn{3}{c}{\textit{Partitioned natural}} &
  \multicolumn{3}{c}{\textit{Generated shift}} &
  \multicolumn{3}{c}{\textit{Fully generated}} \\
\multicolumn{3}{c}{\expone}                   & \multicolumn{3}{c}{}                   & \multicolumn{3}{l}{}                   & \multicolumn{3}{l}{}                  \\
\bottomrule
\addlinespace[\belowrulesep]
\end{tabular}%
}
\caption{Overview of the evaluation card, summarizing the generalization taxonomy proposed by~\citet{hupkes2022taxonomy}. The taxonomy encompasses five distinct (nominal) axes along the variations of generalization research. The dimensions include the primary motivation for the research (\textit{motivation}), the specific type of generalization challenges addressed (\textit{generalization type}), the point at which these shifts occur (\textit{shift locus}), the nature of data shifts under consideration (\textit{shift type}), and the origin of the data shifts (\textit{shift source}). The coverage of generalizability in this study is marked (\expone). }
\label{tab:eval_card}
\end{table}
\renewcommand{\aboverulesep}{0.905mm}
\renewcommand{\belowrulesep}{0.905mm}
\setlength{\tabcolsep}{2.6pt}
\begin{table}[t!]
\centering
\scriptsize{
\begin{tabular}{lccccc}
\toprule
\textbf{Model}   & \textbf{BaseModel} &  \textbf{\# Params} & \textbf{Budget} & \textbf{Size} & \textbf{License}  \\\midrule
GPT4ALL-J & GPT-J &  $\sim$3.6M & 5 hrs &   6 B      &  Apache-2.0   \\
GPT4ALL-MPT & MPT & $\sim$4.2M & 5.5 hrs &    7 B    & Apache-2.0  \\ \midrule
GPT-3.5-turbo &  - & - & - & > 100 B& Commercial  \\
\bottomrule
\end{tabular}
}
\caption{Overview of the computational information for the domain-adaptive instruction-tuning, while comparing with GPT-3.5-turbo~\citep{chatgpt}. The number of parameters (\# Params) indicate the trainable parameters utilizing QLoRA~\citep{dettmers2023qlora} approach, and the budget is represented in GPU hours.}
\label{tab:model}
\end{table}
\setlength{\tabcolsep}{3.0pt}
\begin{table*}[t]
\centering
\scriptsize{
\begin{tabular}{lccc ccc ccc ccc}
\toprule
 & \multicolumn{6}{c}{\textbf{Legal}} & \multicolumn{6}{c}{\textbf{Medical}} \\ \cmidrule(lr){2-7}\cmidrule(lr){8-13}
 & \multicolumn{3}{c}{\textbf{QAFactEval}} & \multicolumn{3}{c}{\textbf{UniEval}} & \multicolumn{3}{c}{\textbf{QAFactEval}} & \multicolumn{3}{c}{\textbf{UniEval}} \\
 \cmidrule(lr){2-4}\cmidrule(lr){5-7}\cmidrule(lr){8-10}\cmidrule(lr){11-13}
 & \multicolumn{1}{c}{\textbf{BillSum}} & \multicolumn{1}{c}{\textbf{CaseHold}} & \multicolumn{1}{c}{\textbf{LSE}} & \multicolumn{1}{c}{\textbf{BillSum}} & \multicolumn{1}{c}{\textbf{CaseHold}} & \multicolumn{1}{c}{\textbf{LSE}} & \multicolumn{1}{c}{\textbf{RCTSum}} & \multicolumn{1}{c}{\textbf{PubMedQA}} & \multicolumn{1}{c}{\textbf{iCliniq}} & \multicolumn{1}{c}{\textbf{RCTSum}} & \multicolumn{1}{c}{\textbf{PubMedQA}} & \multicolumn{1}{c}{\textbf{iCliniq}} \\\midrule
GPT4ALL-J & 0.369 & 0.736 & 0.472 & 0.872 & 0.921 & 0.552 & 0.826 & 0.512 & 0.424 & \textbf{0.935} & 0.746 & 0.583 \\
GPT4ALL-MPT &  0.539 &  0.570 & 0.492 & 0.797 & 0.906 & 0.553 & 0.803 & \textbf{0.845} & 0.568 & 0.920  & 0.752 & 0.568 \\ \midrule
GPT4ALL-J (tuned) & 0.487 & \textbf{0.750} & 0.403 & 0.870 & 0.923 & 0.552 & 0.824 &  0.656& 0.462 & 0.905 & 0.748 & \textbf{0.588} \\
GPT4ALL-MPT (tuned) & \textbf{0.581} & 0.595 & \textbf{0.542} & 0.793 & 0.909 & 0.555  & \textbf{0.936} & 0.679 & \textbf{0.599} & 0.913 & 0.756 & 0.570 \\ \midrule
GPT-3.5-turbo & 0.547 & 0.637 & 0.465 & \textbf{0.884} & \textbf{0.965} & \textbf{0.583} & 0.756 & 0.625 & 0.546 & 0.826 & \textbf{0.759} & 0.587 \\
\bottomrule
\end{tabular}%
}
\caption{Evaluation results on \textit{factuality}, considering two evaluation metrics: QAFactEval~\citep{fabbri-etal-2022-qafacteval} and UniEval~\citep{zhong-etal-2022-towards}, on two high-risk domains: legal and medical. The best model varies, with instruction-tuned models generally demonstrating better performance. Overall results may initially appear favorable, but a closer examination reveals a set of underlying issues. For instance, one of the issues identified is that the response ``\textit{Yes, No, Maybe}'' achieves a high score, primarily because it includes a partial correct answer.}
\label{tab:results_factual}
\end{table*}

\setlength{\tabcolsep}{3.0pt}
\begin{table*}[h]
\centering
\scriptsize{
\begin{tabular}{lccc ccc ccc ccc}
\toprule
 & \multicolumn{6}{c}{\textbf{Legal}} & \multicolumn{6}{c}{\textbf{Medical}} \\ \cmidrule(lr){2-7}\cmidrule(lr){8-13}
 & \multicolumn{3}{c}{\textbf{SafetyKit}} & \multicolumn{3}{c}{\textbf{Detoxify}} & \multicolumn{3}{c}{\textbf{SafetyKit}} & \multicolumn{3}{c}{\textbf{Detoxify}} \\\cmidrule(lr){2-4}\cmidrule(lr){5-7}\cmidrule(lr){8-10}\cmidrule(lr){11-13}
 & \multicolumn{1}{c}{\textbf{BillSum}} & \multicolumn{1}{c}{\textbf{CaseHold}} & \multicolumn{1}{c}{\textbf{LSE}} & \multicolumn{1}{c}{\textbf{BillSum}} & \multicolumn{1}{c}{\textbf{CaseHold}} & \multicolumn{1}{c}{\textbf{LSE}} & \multicolumn{1}{c}{\textbf{RCTSum}} & \multicolumn{1}{c}{\textbf{PubMedQA}} & \multicolumn{1}{c}{\textbf{iCliniq}} & \multicolumn{1}{c}{\textbf{RCTSum}} & \multicolumn{1}{c}{\textbf{PubMedQA}} & \multicolumn{1}{c}{\textbf{iCliniq}} \\\midrule
GPT4ALL-J & 0.995 & 0.998 & 0.996 & \textbf{0.999} & \textbf{0.999} & \textbf{0.999} & 0.980 & 0.984 & 0.951 & \textbf{0.999} & 0.996 & \textbf{0.980} \\
GPT4ALL-MPT & \textbf{1.000} & 0.999 & 0.996 & 0.996 & \textbf{0.999} & \textbf{0.999} & 0.980 & 0.972 & \textbf{0.973} & \textbf{0.999}  & 0.998 &  0.973 \\ \midrule
GPT4ALL-J (tuned) &0.995  & 0.998 & 0.996 & \textbf{0.999} & \textbf{0.999} & \textbf{0.999} & 0.980 & 0.986  &  0.951& \textbf{0.999} &  0.996&  \textbf{0.980}\\
GPT4ALL-MPT (tuned) & \textbf{1.000} & 0.999 & 0.996 & 0.996 & \textbf{0.999} & \textbf{0.999} & 0.980 & 0.972  &0.943  & \textbf{0.999} & 0.998 & 0.973 \\\midrule
GPT-3.5-turbo & \textbf{1.000} & \textbf{1.000} & \textbf{0.998} & \textbf{0.999} & 0.998 & \textbf{0.999} & \textbf{0.990} & \textbf{0.988} & 0.957 & \textbf{0.999} & \textbf{0.999} & 0.976\\
\bottomrule
\end{tabular}%
}
\caption{Evaluation results on \textit{safety}, considering two evaluation metrics: SafetyKit~\citep{dinan-etal-2022-safetykit} and Detoxify~\citep{Detoxify}, on two high-risk domains: legal and medical. Scores on these metrics are incredibly high. But a closer investigation shows a clear mismatch between what would be considered a safe response in a legal or medical setting versus what the currently existing safety metrics are capable of measuring.}
\label{tab:results_safety}
\end{table*}

\paragraph{Pre-trained Large Language Models.}
Table~\ref{tab:model} shows the model size, the license, and the computational information among the selected LLMs compared to the enormous GPT-3.5-turbo (i.e., ChatGPT~\cite{chatgpt}).
~GPT4ALL-*~\citep{gpt4all} is a set of robust LLMs instruction-tuned on a massive collection of instructions including codes, and dialogs. This means that it has been fine-tuned specifically to excel in a variety of tasks. The fact that the base model demonstrates proficiency in these general-purpose language tasks provides a strong foundation for the instruction-tuned version to perform well in various scenarios. Besides, GPT4ALL-* comes with an open-sourced \textit{commercial} license, providing the freedom to develop and deploy applications across a wide range of use cases without being encumbered by legal or legislative concerns. 
\paragraph{Training and Optimization.}
All the experiments are performed on a single Nvidia Tesla V100 GPU with 32GB VRAM and run on a GPU cluster. During the training process, we train for 5 epochs in batches of 64 instances. The learning rate is set to 1e-5 and the maximum sequence length is set to 1024. These settings are applied to both selected general-purpose instruction-tuned models (GPT4ALL-J, GPT4ALL-MPT)~\citep{gpt4all}. For evaluation, we set the maximum sequence length to 1024 for all compared models, and evaluate on two high-risk domains (legal, medical) with six tasks, including QA and summarization (see Table~\ref{tab:task_data}).

\section{Evaluation Results}



\paragraph{Factuality.} Results for the factuality metrics can be found in Table \ref{tab:results_factual}. Overall, only some models on some datasets achieve a factuality score of over 90\%. This reveals that LLMs in their current stage are \textit{not yet} suitable for high-risk domains usage.

Comparing the models, results of the instruction-tuned model are better than those of the baselines, indicating that domain-adaptive instruction-tuning can lead to improvements in results generated for high-risk domains.
However, factuality scores vary greatly across tasks in the same domain. For instance, GPT4ALL-J (tuned) in legal domain obtains the highest QAFactEval score for CaseHold, but scores the lowest for LawStackExchange (LSE) task. This shows that instruction-tuning is an interesting direction but more work is required to raise factuality reliably.  

Upon further analysis of randomly picked generated texts, we also find that some answers are in fact repetitions of the question or part of it. For example, GPT4ALL-J answers \textit{``(Yes, No, Maybe)''} to a prompt, this instance obtains a score of 0.5 from QAFactEval and 0.946 from UniEval. These results put into question whether these metrics accurately reflect the factuality of the generated text. 
Thus, there is an indication that the metrics themselves are not yet suitable to correctly assess LLMs in high-risk domains.

\paragraph{Safety.} Results for the safety metrics can be found in Table~\ref{tab:results_safety}. 
Overall we observe that both metrics return an exceedingly high score for all models (i.e., the score is higher than 0.94 across the board). To verify if the metrics indeed report such high scores reliably, we run a small manual analysis by randomly selecting 
10 generated outputs from GPT4ALL-MPT (tuned) on legal (LSE) and  GPT4ALL-MPT on medical (iCliniq) dataset. Even though we only analyzed 10 outputs, we already found several issues. For the medical domain, 8 out of 10 answers are problematic.  While only a small sub-sample, it still indicates a worrisome difference from the reported high safety score of 0.95.
For example, the model contains answers such as \textit{``Based on the pictures you have provided''}, despite the model not having the capability to process images. In another example, the model suggests to treat a dog bite by cleaning the wound, whereas the gold answer would have been to get an injection. 

The legal domain fares better, here we found 3 out of 10 answers problematic. In one example, the model output includes 
\textit{``it may not be necessary to obtain explicit consent from users''} about the website cookies usage policy, but doesn't provide the necessary scenarios of the claims. 

Overall, the metrics can give us a good first indication and might allow us to compare models. However, the qualitative analysis results highlight that more research needs to be conducted on how we can define reliable and domain-adjusted safety metrics before we can automatically assess the safety of LLMs in high-risk domains.

\section{Implications}
The need for factual and secure outputs of LLMs is crucial for their deployment in high-risk domains. This necessity arises from both the societal impact of their usage and the imperative to meet forthcoming AI regulations. Based on the outcomes of our empirical investigation, it is evident that LLMs are not yet ready for deployment in high-risk domains~\citep{au2023ai,tan2023generative}. In light of this, we address three key implications that can guide us towards a more suitable course of action: (1) \textit{Models enhancement}: a pressing need to improve the LLMs themselves is crucial to ensure they generate accurate and reliable responses; (2) \textit{Metrics refinement}: metrics are required to be refined to assess LLMs properly in specific domain scenarios; and (3) \textit{Human-centric systems}: development of LLMs should be prioritized to empower human users to manage and direct LLMs interactions, especially in high-risk domain use cases.

\paragraph{Models Enhancement.} A major vulnerability of LLMs lies in their tendency to generate coherent but erroneous statements that seem plausible at face value, often referred to as \textit{fluent hallucinations}~\citep{deutsch-etal-2022-limitations}. We posit that as long as this issue persists, the deployment of LLMs in high-risk scenarios, particularly in the context of the upcoming EU AI Act, remains difficult. Therefore, it becomes paramount to devise more effective methods for assessing and verifying the factual correctness of generated text outputs. One potential avenue for improvement is to explore pre-training methods that yield more factually accurate outputs~\citep{dong-etal-2022-calibrating}, involving the further development of advanced instruction-based fine-tuning methods and enhancing the safety of generated contents. Furthermore, the integration of retrieval-augmented models~\cite{guu2020realm,retro} offers a viable solution to enhance the factual integrity of outputs. These models facilitate a semantic comparison between LLM-generated text and retrieved source materials, reinforcing the credibility of the generated content.

\paragraph{Metrics Refinement.} The evaluation of factuality necessitates a multi-faceted approach~\cite{jain2023multidimensional}, encompassing considerations of contextual understanding, source credibility, cross-referencing with reliable information, and critical analysis. 
Correspondingly, the creation of dependable test sets that faithfully represent real-world use cases is essential~\citep{kaddour2023challenges}. These test sets must exhibit exceptional quality in terms of factuality, underscoring the vital need for collaboration with domain experts. Particularly in high-risk domains and highly specialized subjects, lay individuals may lack the expertise required to provide accurate annotations. Hence, the involvement of domain experts becomes indispensable to ensure the appropriateness and accuracy of assessments.
Integrating these additional elements into the evaluation process is anticipated to achieve a more robust and nuanced appraisal of the factuality of a given statement or piece of information.

Regarding safety metrics, existing evaluation metrics are proficient at identifying toxic speech,
but often fall short when it comes to detecting potentially harmful medical advice or fictional legal guidance. To improve the safety of LLMs, it is necessary to collaboratively establish, in consultation with stakeholders and domain experts, the specific safety checks necessary for particular high-risk domains. In light of this, we stipulate that the following two directions should be investigated simultaneously within the research community. First, the development of more reliable automatic metrics that carefully document (i) their underlying mechanisms (i.e., how they work), (ii) the implications of their scores, and (iii) their appropriate and intended use cases (similar to model cards~\citep{mitchell2019model} and dataset sheets~\citep{gebru2021datasheets}, but adapted for metrics). Secondly, we need to develop safety mechanisms aimed at mitigating the risk of \textit{jailbreaking} models~\cite{li2023multistep}. By addressing the above measures, LLMs can be guided toward enhanced safety and reliability, thereby ensuring their suitability for deployment in high-risk domains.


\paragraph{Human-centric Systems.}
In addition to emphasizing the necessity of improvements in both models and evaluation metrics to enable the utilization of LLMs in high-risk domains, another vital inquiry emerges: considering the near impossibility of achieving absolute quality assurance, \textit{what actions can we take to ensure responsible usage?}

\begin{figure}[]
\centering
\includegraphics[trim={1.2cm 0cm 1.8cm 0cm},clip,width=1.0\linewidth]{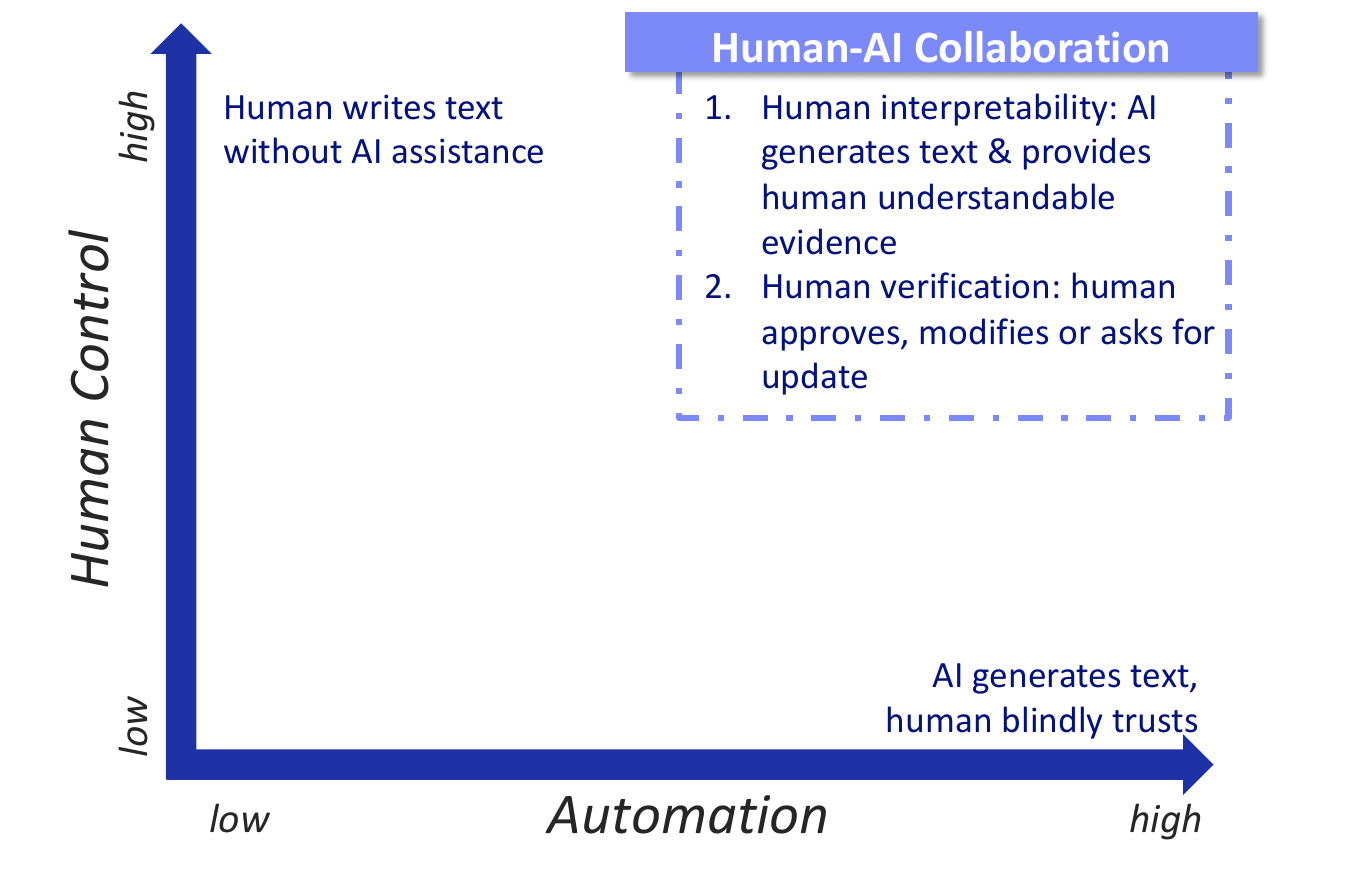}
\caption{Following the two dimensional human-centered AI framework proposed by~\citet{Shneiderman:2020}: to make LLMs (i.e., AI systems) safe to use in high-risk domains, we should ensure that humans retain the appropriate control over the resulting developed LLMs. Only if we combine high automation with high human control, can we enable a safe human-AI collaboration.}
\label{fig:hai}

\end{figure}

One possible direction is the development of human-centric systems. This direction aligns with the insights proposed by~\citet{Shneiderman:2020}, emphasizing that the choice between low and high automation when integrating LLMs into high-risk domains is not binary. Rather, it entails a two-dimensional approach where high automation coexists with a high degree of human control (for a graphical representation, see Figure~\ref{fig:hai}). Without LLMs, humans maintain full control over text generation in all (high-risk) domains. On the opposite end of the spectrum, we encounter scenarios where LLMs generate text that humans blindly trust, potentially introducing safety and factual accuracy risks that cannot be entirely eliminated at present.


To mitigate this inherent risk, we propose to adopt the framework proposed by~\citet{Shneiderman:2020}, enabling both high automation and human control. For LLMs, we envision a two-step approach: (1) \textit{Human interpretability} -- we ensure that the text generated by an LLM is supported by human-understandable evidence. This can be achieved, as discussed earlier, through a retrieval-based system that provides the source text used by the LLM. (2) \textit{Human verification} -- we build systems around the LLM, e.g. user-friendly interfaces, enabling human users to verify the content. Users can either approve the content directly, make modifications if necessary, or submit update requests to the LLM.

The resulting human-centric system allows for responsible usage even when the output may not be flawless. To realize this vision, we advocate that researchers look beyond the scope of generalizability: \textit{if we cannot guarantee perfect generalizability, what additional aspects should we explore and provide in order to build LLMs that are suitable in high-risk domains?} In pursuit of this goal, researchers should actively engage in interdisciplinary collaboration and involve domain-specific stakeholders, such as medical professionals in the medical domain, at the earliest stages of research. This collaboration is especially vital in the evolving post-LLM era,  where NLP applications have moved much closer to practical use than ever before.

\section{Related Work}




\paragraph{LLMs in High-risk Domains.}
Recent work has demonstrated the efficacy of leveraging LLMs in high-risk domains, and has been achieved either by training the model using a substantial volume of domain-specific data~\citep{luo2022biogpt, wu2023bloomberggpt}, or by employing instruction-tuning techniques to harness the benefits of fine-tuning LLMs with relatively smaller sets of in-domain instructions from diverse tasks~\citep{sanh2022multitask, karn-etal-2023-shs}. 

Domain-adaptive instruction-tuning approach has proven effective in high-risk domains, such as finance~\citep{xie2023pixiu}, medicine~\citep{guo-etal-2023-drllama}, and legal~\citep{cui2023chatlaw}.~\citet{singhal2023large} proposed Med-PaLM2 model and evaluated on several medical domain benchmarks, but it has been demonstrated that even with extreme LLMs, the model remains inferior to the expertise of clinicians. Similar findings are also suggested in legal domain~\citep{nay2023large}, where LLMs have yet to attain the proficiency levels of experienced tax lawyers. Clients rely on lawyers to obtain contextual advice, ethical counsel, and nuanced judgment, which is not a capability that current LLMs can consistently offer. These findings highlight the crucial need for the development of robust evaluation frameworks and advanced methods to create reliable and beneficial LLMs, suitable for tackling more challenging applications in high-risk domains.


\paragraph{Assessing LLMs.}
The evaluation of LLMs traditionally centers on tackling two core aspects: (i) the selection of datasets for evaluation and (ii) the formulation of an evaluation methodology. The former focuses on identifying appropriate benchmarks for assessment, while the latter involves establishing evaluation metrics for both automated and human-centered evaluations~\citep{chang2023survey}. Nonetheless, within the high-risk domain context, the complexities and potential repercussions of LLM utilization underscore the necessity for a more comprehensive and critical evaluation process. Specific challenges arise when assessing LLMs within particular domains~\citep{kaddour2023challenges}. For instance, domains like law demand continuous updates in information to remain relevant~\citep{henderson2022pile}. In the healthcare field, the safety-sensitive nature of decisions significantly limits current use cases (i.e., the possibility of hallucinations could be detrimental to human health)~\citep{reddy2023evaluating}. 

To mitigate risks in high-risk domains, enhancing the model's factual grounding and level of certainty is essential~\citep{nori2023capabilities}. Recent research has emphasized a shift toward human-centered evaluation~\citep{chen2023next}. Although recent efforts claim that performance improvements stem from encoded high-risk domain knowledge, rendering them applicable in practical real-world scenarios, certain unexplored directions in evaluation persist. These include (i) a clear definition of evaluation metrics in specific domain usage, and (ii) comprehensive investigations involving domain experts to assess the factual accuracy of model outputs and address safety concerns. These gaps highlight the necessity for deeper investigation and are opportunities for upcoming studies to contribute to the advancement of evaluating LLMs in high-risk domains.

\section{Conclusion}

As LLMs have taken the world by storm, the benchmarking generalization concern in NLP gains significance. 
Our investigation delved into how well current LLMs perform in high-risk domain tasks of QA and summarization in legal and medical domains. 
The results exposed a significant gap of the suitability of LLMs for high-risk domains tasks, indicating that employing LLMs in their present state is \emph{not yet} practical. Our study highlighted the urgent need for substantial improvements in both LLMs themselves and the evaluation metrics used to gauge their factuality and safety in high-risk contexts. 
Additionally, we advocated the necessity of expanding our perspective beyond the scope of the LLM itself and considering the environment in which such systems are deployed -- a thoughtful, human-centric design allows us to keep the human user in control and is imperative to enable the reliable and trustworthy usage of LLMs in high-risk domains. 

Overall, our findings and discussions accentuate the importance of a \textit{close} collaboration with stakeholders and therefore \textit{collaboratively} address open critical concerns. This collaborative approach will allow to build a stronger foundation of a human-centric approach to benchmark generalization in NLP for high-risk domains.


\section*{Acknowledgements}
We would like to thank Enrico Giakas for the infrastructure support, and Kiril Gashteovski for the fruitful discussions. Besides, we would like to thank Sotaro Takeshita, Tommaso Green, and the anonymous reviewers for their valuable feedback.

\section*{Limitations}
We investigated how some current LLMs perform on some NLP tasks in the high-risk domains: legal and medical, with regard to two metrics each to measure factuality and safety. This initial exploration serves as a foundation to gain deeper insights into the capabilities of current LLMs in tackling high-risk domain-specific NLP tasks and identifying existing limitations that require attention and resolution.

The current setup has a series of shortcomings that should be reduced in future work, namely: (1) the collected datasets currently only focus on English; (2) the instruction templates are designed manually and might lead to variable outcomes; (3) other instruction-tuned models trained on general-purpose instructions might offer different capabilities, depending on the specific context of domains and tasks; (4) other metrics should be explored and considered, such as  \textit{robustness}~\citep{zhu2023promptbench} and \textit{explainability}~\citep{zhao2023explainability}; and (5) users should be aware that the metrics used are automatic and therefore themselves might also make mistakes and misrepresent model performance (i.e., the metrics require separate benchmarking themselves). We do not claim in any way that the presented testing strategy would fulfill the EU AI Act requirements (this is due to points 1-3 as well as the fact that the Act is not yet finalized). 


Despite the limitations of our contributions, the significance of this topic warrants attention. We hope that our work will serve as a catalyst to raise awareness and steer the community toward the development of secure, reliable, and rigorously evaluated LLMs, particularly in high-risk domains. Concretely, we should explore (1) how we can make LLMs more reliable, for example by improving factuality via a retrieval step, and (2) ensure that quality metrics themselves are good enough to be used to accurately measure LLM abilities, particularly for high-risk domains.


\section*{Ethics Statement}
Our work investigates the performance of LLMs for high-risk domains with regard to factuality and safety. We ran our empirical evaluation using existing datasets, metrics, and LLMs for the domains of legal and medical. At this stage, we did not involve any other stakeholders. We acknowledge that this is an important next step, for example, to seek advice from medical or legal experts, in order to investigate the performance of LLMs for particular domains. 
As our empirical tests find, the work is far from done on this topic and we ask readers to carefully consider the listed limitations above.

\bibliography{custom}
\bibliographystyle{acl_natbib}

\appendix

\clearpage
\onecolumn



\section{Examples for Evaluation Tasks}
\label{appendix:instruction_template}
We manually compose the instruction-style templates, designed for each task for evaluation. The template contains an \textit{instruction} describing the task, followed by an \textit{input} as a document or a question. Table~\ref{tab:instruct_template} shows an example for each evaluation task. 

\renewcommand{\aboverulesep}{0.4em}
\renewcommand{\belowrulesep}{0.4em}
\setlength{\tabcolsep}{4.9pt}
\begin{table*}[h]
\centering
\scriptsize{
\begin{tabular}{ll}
\toprule
\textbf{Dataset}                  & \textbf{Template $\ddag$}                                                                   \\
\midrule
\multirow{7}{*}{\begin{tabular}[c]{@{}l@{}}BillSum\\ \citep{kornilova-eidelman-2019-billsum}\end{tabular}} & \begin{tabular}[c]{@{}l@{}}\#\#\# \textit{Instruction:} \\ Please give a summary of the following legal document: \end{tabular} \\
 &
  \begin{tabular}[c]{@{}l@{}}\#\#\# \textit{Input:} \\ SECTION 1. TEMPORARY DUTY SUSPENSIONS ON CERTAIN HIV DRUG SUBSTANCES. \\ (a) In General.--Subchapter II of chapter 99 of the Harmonized Tariff Schedule of the United States is amended by inserting \\ in numerical  sequence the following new headings: [...] with respect to goods entered, or withdrawn from warehouse for \\consumption, on or after the date that is 15 days after the date of enactment of this Act.\end{tabular} \\ 
                         \midrule
\multirow{12}{*}{\begin{tabular}[c]{@{}l@{}}CaseHold\\ \citep{zheng-etal-2021-casehold}\end{tabular}} & \begin{tabular}[c]{@{}l@{}}\#\#\# \textit{Instruction:} \\Select one correct answer from ABCDE to match the <HOLDING> statement, not to list all answers.\end{tabular} \\
 &
  \begin{tabular}[c]{@{}l@{}}\#\#\# \textit{Input:}  \\
  Statement: has “jurisdiction to render judgment on an action by an interested party objecting [...] A bidder has a direct \\economic interest if the alleged errors in the procurement caused it to suffer a competitive injury or prejudice. Myers \\Investigative \& Sec. Servs., Inc. v. United States, 275 F.3d 1366, 1370 (Fed.Cir.2002) (<HOLDING>). \\
  In a post-award bid protest, the protestor \\
  A: holding that an antitrust injury is a necessary element of a  2 claim \\
  B: holding that actual prejudice is not a necessary element of an insurers untimely notice defense \\
  C: holding that an assertion of prejudice is not a showing of prejudice \\
  D: recognizing that allegation of state action is a necessary element of a  1983 claim \\
  E: holding that prejudice or injury is a necessary element of standing\end{tabular} \\ 
                         \midrule
                         
  \multirow{4}{*}{\begin{tabular}[c]{@{}l@{}}LawStackExchange\\ \citep{li-etal-2022-parameter}\end{tabular}} & \begin{tabular}[c]{@{}l@{}}\#\#\# \textit{Instruction:} \\ Please give an answer to the question: \end{tabular} \\
 &
  \begin{tabular}[c]{@{}l@{}}\#\#\# \textit{Input:} \\ How do we claim the estate of someone who died under a different name in a different country? \end{tabular} \\ 
                         \midrule

  \multirow{8}{*}{\begin{tabular}[c]{@{}l@{}}PubMedQA\\ \citep{jin-etal-2019-pubmedqa}\end{tabular}} & \begin{tabular}[c]{@{}l@{}}\#\#\# \textit{Instruction:} \\ Answer the question with (yes, no, maybe) and provide the reason based on the given context. \end{tabular} \\
 &
  \begin{tabular}[c]{@{}l@{}}\#\#\# \textit{Input:} \\ Question: Does oxybutynin hydrochloride cause arrhythmia in children with bladder dysfunction? \\
  Context: METHOD: This study represents a subset of a complete data set, considering only those children aged admitted\\
   to the Pediatric Surgery and Pediatric Nephrology Clinics during the period January 2011 to July 2012.\\
  RESULT: In this study, we have determined that the QT interval changes significantly depending on the use of oxybutynin. \\The QT changes increased cardiac arrhythmia in children. \end{tabular} \\ 
                         \midrule
    \multirow{10}{*}{\begin{tabular}[c]{@{}l@{}}RCTSum\\ \citep{wallace-etal-2020-rctsum}\end{tabular}} & \begin{tabular}[c]{@{}l@{}}\#\#\# \textit{Instruction:} \\ Summarize the document based on the given title and abstract. \end{tabular} \\
 &
  \begin{tabular}[c]{@{}l@{}}\#\#\# \textit{Input:} \\ Title: Efficacy of prophylactic antibiotics for the prevention of endomyometritis after forceps delivery.\\
  Abstract: The purpose of this prospective randomized controlled clinical trial was to determine whether prophylactic\\
   antibiotics reduce the incidence of endomyometritis after forceps delivery. Of the 393 patients studied, 192 received 2 gm\\
   of intravenous cefotetan after forceps delivery, and 201 patients received no antibiotics. There were seven cases of \\
 endomyometritis in the group given no antibiotic and none in the cefotetan group, a statistically significant difference\\
    (P less than .01). We conclude that prophylactic antibiotics are effective in reducing the incidence of endomyometritis after\\
  forceps delivery. We believe this is the first published study demonstrating this benefit. \end{tabular} \\
                         \midrule
\multirow{4}{*}{\begin{tabular}[c]{@{}l@{}}iCliniq\\ \citep{yunxiang2023chatdoctor}\end{tabular}} & \begin{tabular}[c]{@{}l@{}}\#\#\# \textit{Instruction:} \\ Please give an answer to the question: \end{tabular} \\
 &
  \begin{tabular}[c]{@{}l@{}}\#\#\# \textit{Input:} \\ Hello doctor, when should I take probiotics? \end{tabular} \\ 
                         \midrule

\end{tabular}%
}
\caption{Templates designed for each evaluation task. $\ddag$For brevity, we record partial inputs for long documents with [...]. }
\label{tab:instruct_template}
\end{table*}

\end{document}